\documentclass[letterpaper]{article} 
\usepackage[]{aaai23}  
\usepackage{times}  
\usepackage{helvet}  
\usepackage{courier}  
\usepackage[hyphens]{url}  
\usepackage{graphicx} 
\usepackage{natbib}  
\usepackage{caption} 
\usepackage{algorithm}
\usepackage{algorithmic}

\usepackage{newfloat}
\usepackage{listings}
\DeclareCaptionStyle{ruled}{labelfont=normalfont,labelsep=colon,strut=off} 
\floatstyle{ruled}
\newfloat{listing}{tb}{lst}{}
\floatname{listing}{Listing}
\usepackage{xcolor}

\usepackage{amsmath,amsfonts,bbm}
\usepackage{multirow}
\usepackage{booktabs}

\title{Mulco: Recognizing Chinese Nested Named Entities Through Multiple Scopes}

\author{
    Jiuding Yang\equalcontrib\textsuperscript{\rm 1},
    Jinwen Luo\equalcontrib\textsuperscript{\rm 2},
    Weidong Guo\equalcontrib\textsuperscript{\rm 2},
    Jerry Chen\textsuperscript{\rm 1},
    Di Niu\textsuperscript{\rm 1},
    Yu Xu\textsuperscript{\rm 2}
}
\affiliations{
    \textsuperscript{\rm 1}University of Alberta\\
    \textsuperscript{\rm 2}Platform and Content Group, Tencent\\

}

\begin{document}
\maketitle
\begin{abstract}
Nested Named Entity Recognition (NNER) has been a long-term challenge to researchers as an important sub-area of Named Entity Recognition. NNER is where one entity may be part of a longer entity, and this may happen on multiple levels, as the term nested suggests. These nested structures make traditional sequence labeling methods unable to properly recognize all entities. While recent researches focus on designing better recognition methods for NNER in a variety of languages, the Chinese NNER (CNNER) still lacks attention, where a free-for-access, CNNER-specialized benchmark is absent. In this paper, we aim to solve CNNER problems by providing a Chinese dataset and a learning-based model to tackle the issue. To facilitate the research on this task, we release ChiNesE, a CNNER dataset with 20,000 sentences sampled from online passages of multiple domains, containing 117,284 entities failing in 10 categories, where 43.8 percent of those entities are nested. Based on ChiNesE, we propose Mulco, a novel method that can recognize named entities in nested structures through multiple scopes.
Each scope use a designed scope-based sequence labeling method, which predicts an anchor and the length of a named entity to recognize it. Experiment results show that Mulco has outperformed several baseline methods with the different recognizing schemes on ChiNesE. We also conduct extensive experiments on ACE2005 Chinese corpus, where Mulco has achieved the best performance compared with the baseline methods.
\end{abstract}
\section{Introduction}
\label{sec-intro}

As a classic and popular topic of Natural Language Processing (NLP), Named Entity Recognition (NER) has been widely studied to benefit real-world NLP applications (e.g., recommendation systems). By defining NER as a sequence labeling task, previous research has achieved great performance on the recognition by employing neural-based architectures \cite{10.1109/TKDE.2020.2981314}. 

Nested NER (NNER) has raised new challenges for researchers because of the deeply nested structures of entities \cite{wang2022nested}.
Due to the natural properties of languages, the tokens of an entity may be within other larger entities. 
For example, ``Beijing", as an entity of location, may be included in "Beijing Municipal Government" and ``the spokesman of Beijing Municipal Government". Those two examples are entities of organization and person respectively. 
Such nested structures bring great challenges to recognition, which also commonly exist in Chinese Nested Named Entity Recognition (CNNER). As a sub-field of NNER, CNNER plays a key role in many downstream tasks (e.g., entity linking, information retrieval) and real-world applications (e.g., recommendation, search engines). 
While many English NNER (ENNER) datasets are released to support the research in such a field, there exist few available datasets on CNNER, which impedes the advancement of CNNER research. Existing CNNER dataset either has access problems \cite{doddington-etal-2004-automatic} or contains a low ratio of nested named entities \cite{yu2018corpus}. Also, they usually do not have an official parsing method and test set which has caused non-unified metrics \cite{xue2002building, doddington-etal-2004-automatic, yu2018corpus}.

In this paper, to facilitate the research on CNNER, we construct ChiNesE, a \textbf{Chi}nese \textbf{Nes}ted Named \textbf{E}ntity Recognition dataset. It consists of 20,000 sentences sampled from various domains, containing 117,284 entities falling into 10 major categories. In ChiNesE, 43.8\% of all entities are in nested structures, with a maximum depth level of 8. To the best of our knowledge, ChiNesE is currently the largest (number of nested named entities) fully open-sourced CNNER dataset with official data split for a unified metric.

In order to tackle the challenges brought by NNER, various options have been developed to extract English named entities from nested structures, which can also be re-implemented to solve CNNER. Since traditional NER methods such as ``IOB'' and ``BIOES'' \cite{10.1109/TKDE.2020.2981314} fail to encode nested named entities, researchers have tried to re-enable the sequence labeling method by splitting NNER into multiple sub-tasks. They then predict all corresponding sequences of the sub-labels \cite{alex2007recognising}, or stack all sub-labels of a word into one joint-label and recognize nested named entities by predicting such labels \cite{6234172}. However, most of these approaches focus on solving NNER with a limited depth of nested structures, which does not satisfy the need of today's applications.

Realizing the limitation of the sequence labeling method, researchers tend to solve the NNER task with different labeling schemes, and multiple new research directions have been developed. For example, 
the span-based \cite{yu-etal-2020-named, tan2020boundary, li2022unified} methods recognize named entities from text spans of the input text. They need to aggregate the span information before the classification, which greatly increases the complexity of their models.
Transition-based methods \cite{wang-etal-2018-neural-transition, marinho2019hierarchical} construct a binary tree out of the input text, and predict a sequence of actions to reconstruct the tree. However, the error of the action prediction will be accumulated since it is generated sequentially, especially in CNNER, where entities may consist of many characters rather than a few words in ENNER.
Although these recent methods have proven to enhance the performance of NNER, most of them are not tested on CNNER. Moreover, they primarily attempt to solve NNER with new labeling approaches, rather than exploring the potential of applying the sequence labeling approach to NNER.

To fill in the gaps above, based on ChiNesE, we revisit sequence labeling and propose Mulco, a novel method that recognizes Chinese nested named entities through four different scopes. Our proposed method utilizes a scope-based sequence labeling scheme, which uses an anchor and the length of an entity to encode its position and category. Such a sequence labeling method encodes entities in the unit of individual characters, and is thus much easier to learn by models.
By combining multiple scopes using straightforward model structures, the Mulco can overcome the limit that traditional sequence labeling methods suffer from on CNNER tasks and effectively recognizes the named entities from nested structures.

To evaluate the performance of Mulco, we further parse the Chinese corpus of ACE2005 following the existing parsing method of ACE2005 English NNER, and reproduce a number of single-model baseline methods with different labeling schemes on both ChiNesE and ACE2005 for comparison, including Pyramid \cite{wang-etal-2020-pyramid} and Biaffine \cite{yu-etal-2020-named}.
Experimental results show that Mulco outperforms all single-model baselines on both datasets. Our code is released at \textit{Anonymous}, together with the full ChiNesE and the parsing script of ACE2005 Chinese corpus to facilitate future research on CNNER.

\section{Related Works}
\textbf{NER.}
Traditional Named Entity Recognition (NER) methods mainly focus on flat NER, where no named entity is overlapped with the others. Early research was mostly based on rule-based methods \cite{rau1991extracting, mikheev1999named}. Later the development of neural networks draws research attention to combining feature engineering with machine learning \cite{petasis2001using, whitelaw2003evaluating}, where NER is usually defined as a sequence labeling task.
The increasing availability of computing resource further push the research direction more to utilize deep learning models \cite{10.1109/TKDE.2020.2981314}, where pre-trained word (character) vectors \cite{pennington2014glove} and pre-trained language models \cite{devlin-etal-2019-bert} are commonly used to facilitate such works.

\textbf{NNER.}
Nested Named Entity Recognition (NNER) is an important sub-area of NER, where an entity may contain other entities or be a part of other entities \cite{wang2022nested}. Same as NER, early methods address NNER with rule-based methods \cite{houfeng2005simple,8200102}. Later works on NNER combine the pre-trained language models and word vectors with deep learning method to for better performance, which can be mainly classified into four steams: layer-based \cite{wang-etal-2020-pyramid,shibuya2020nested}, span-based \cite{BA1,BA2, yu-etal-2020-named}, transition-based \cite{wang-etal-2018-neural-transition,marinho2019hierarchical} and hypergraph-based \cite{katiyar-cardie-2018-nested, wang2018neural}.
To named a few, \citet{wang-etal-2020-pyramid} designed a layer-based method called Pyramid. It stacks flat NER layers where each layer is utilized to recognize text pieces in different length. 
\citet{yu-etal-2020-named} treat NER as a parsing problem, and develops the span-based method utilizing biaffine attention to address the possibility of each text span as a mention. 
\citet{wang-etal-2018-neural-transition} designed a transition-based method which labels each sentence into a tree structure and tackle NNER problem by predicting a sequence of actions to reconstruct the trees. 
\citet{katiyar-cardie-2018-nested} constructs a hypergraph structure out of a given sentence based on ``BIOES'' sequence labeling tags, and use a standard LSTM-based sequence labeling model to learn the nested entity hypergraph structure. 
There are also other NNER research using generative approach \cite{li-etal-2020-unified}, or use head words as anchors to facilitate the classification of named entities \cite{lin-etal-2019-sequence}. 

\textbf{CNNER.} Compared with English NNER, Chinese NNER (CNNER) has much less research attention \cite{zhang2008fusion, hao2013product, li2021bridge}. Here we introduce some recent works of CNNER.
\citet{6234172} develop their method based on the assumption that the nested depth of most Chinese named entities is no deeper than two. They reformulate CNNER as a cascaded chunking problem on a sequence of words. However, their method requires pre-segmentation of a Chinese sentence, while in most cases such segmentation is not provided. Also, their method can not recognize entities with a depth greater than two.
\citet{BA1,BA2} propose region-based methods, which first detect boundaries of NEs, then assemble them into candidates for further recognition. They use multiple steps to recognize named entities, which increases the inference time.
\citet{li2022unified} propose a unified NER model which considers CNNER. It is a span-based method that encodes the entities into word-to-word relations and predicts the relations between all words in the text to extract the nested named entities. It fuses the predictions of an MLP predictor and a Biaffine \cite{yu-etal-2020-named} predictor, forming an ensemble model that requires much more resources to train compared to other single-model methods.

\begin{table*}
	\centering
	\small
	\begin{tabular}{l c c c c c c c c}
	\toprule 
	\multirow{2}{*}{}&\multicolumn{4}{c}{ACE 2005}&\multicolumn{4}{c}{ChiNesE}\\\cmidrule(l){2-5}\cmidrule(l){6-9} 
	& train & valid  & test &all & train & valid  & test &all\\
	\midrule
	    \#sentence
	    &5,999&727&734&7,460
	    &175,00&1,000&1,500&20,000\\
	    \ \ \ \ \#nested
	    &3,033&387&430&3,850
	    &7,500&990&1,484&9,974\\
	    \midrule
	    \#entity
	    &27,590&3,316&3,812&34,727
	    &95,151&8,833&13,300&117,284\\
	    \ \ \ \ \#nested
	    &12,464&1,480&1,801&15,745
	    &36,289&6,075&9,045&51,409\\\midrule
	    avg. char.
	    &42.9&40.8&45.3&42.9
	    &60.8&65.9&66.7&61.5\\\midrule
	    max depth
	    &\multicolumn{4}{c}{7}&\multicolumn{4}{c}{8}\\\midrule
	    \#category
	    &\multicolumn{4}{c}{7}&\multicolumn{4}{c}{10}\\
	    \bottomrule 
	\end{tabular}
	\caption{The statics of ACE 2005 and ChiNesE. 43.8\% of the entities in ChiNesE are nested , and 45.3\% of entities in ACE 2005 are nested. ``avg. char.'' represents the average characters of the sentences in each dataset.}
	\label{tab-static}
\end{table*}

\textbf{CNNER Datasets.} Many early CNNER research \cite{6234172,zhang2014parsing} is developed based on the Chinese corpus provided by People's Daily 1998 \cite{yu2018corpus} and Chinese Tree Bank (CTB) \cite{xue2002building}. However, People's Daily only has three categories for named entities, and the number of the nested named entities and their depth is much lower than existing English NNER datasets \cite{ringland2019nne, doddington-etal-2004-automatic}. CTB is mainly used for Chinese parsing, thus research developed based on it usually requires parsing information to recognize named entities \cite{zhang2014parsing}.
Recent works \cite{BA1,BA2,li2022unified} parse ACE 2004 and ACE 2005 Chinese corpus \cite{doddington-etal-2004-automatic} and collects a CNNER dataset to develop their methods. However, the detailed parsing and the data split methods are implicit, which makes later researchers hard to follow their works. Moreover, the ACE corpus is not free for access, which could cause difficulties for researchers to study CNNER. 



\section{Dataset}

\begin{figure*}[t]
\centering
\includegraphics[width=0.9\textwidth]{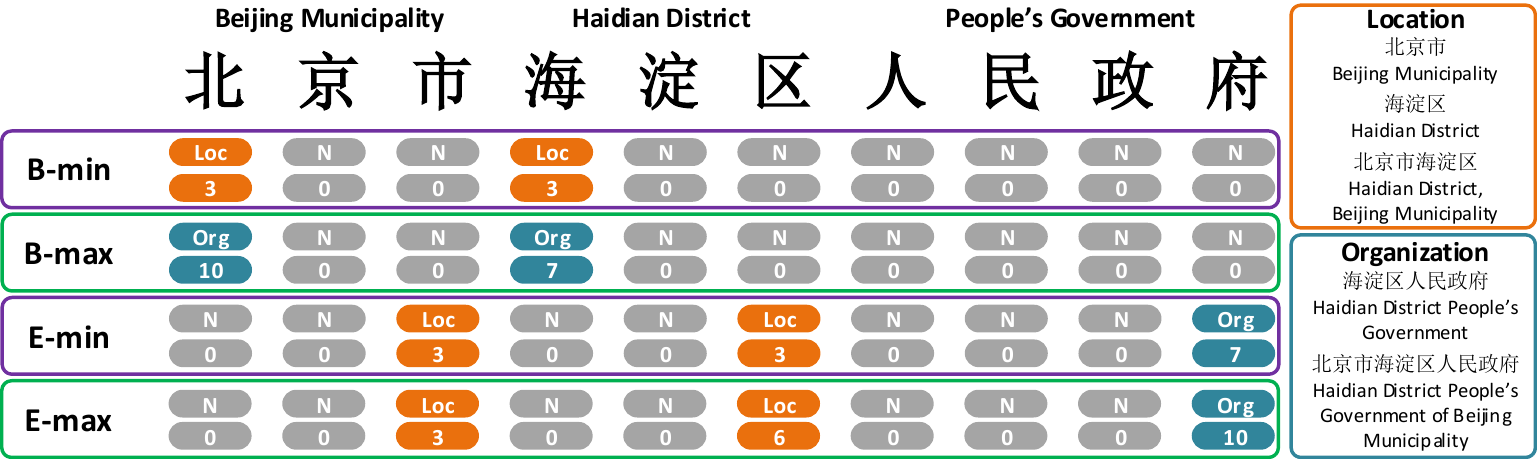} 
\caption{An example of recognizing nested named entities in ``Haidian District People’s Government of Beijing Municipality''.}
\label{fig-example}
\end{figure*}

\begin{table}
\small
\centering
\begin{tabular}{lr}
\hline
\textbf{ACE 2005} & \textbf{Amount}\\
\hline
Person & 14,539 \\
Location & 1,550 \\
Organization & 6,886 \\
Geo-Political & 9,033 \\
Facilities & 1,662 \\
Vehicle & 683 \\
Weapon & 374 \\
-\\
-\\
-\\
\hline
\end{tabular}
\begin{tabular}{lr}
\hline
\textbf{ChiNesE} & \textbf{Amount}\\
\hline
Person & 20,510 \\
Location & 32,924 \\
Organization & 20,194 \\
Time & 11,583 \\
Work & 11,014 \\
Food & 5,647 \\
Product & 4,352 \\
Medicine & 3,927 \\
Event & 3,807 \\
Creature & 3,326 \\
\hline
\end{tabular}
\caption{The distribution of the entities in ACE 2005 and ChiNesE.}
\label{tab-distribution}
\end{table}
We construct ChiNesE to facilitate the research on \textbf{Chi}nese \textbf{Nes}ted Named \textbf{E}ntity Recognition. Given a sentence, CNNER aims to recognize all existing named entities in the nested structures. Figure~\ref{fig-example} is an example of how a nested structure contains multiple nested named entities. Detailed instruction to NNER is given by \citet{wang2022nested}

\textbf{Construction of categories.} The categories of ChiNesE are selected from Chinese Wikipedia \footnote{https://dumps.wikimedia.org/zhwiki/}. We collect the tag of all entities and keep those of the highest frequency. We then manually filtered out the labels that have redundant meaning from the other labels and cut off the labels that are not familiar to most people. After that, we classified the labels into 10 major categories.

\textbf{Data collection and annotation.} The annotation of nested named entities is very hard and expensive.
To control the expense and ensure the quality, we construct ChiNesE in three steps. 

First, we collect 5,000 online passages and use TextRank4ZH\footnote{https://github.com/letiantian/TextRank4ZH} to extract their key sentences. After that, three annotators with professional knowledge backgrounds are hired to label all named entities in the sentences. To ensure the quality, for each sentence, we let two annotators find named entities and one annotator to check the results.

Then, to collect more samples, we train a Biaffine NER model \cite{yu-etal-2020-named} with BERT \cite{devlin-etal-2019-bert} encoder on the annotated dataset of the first step. We run the model on another 45,000 sentences extracted from the different passages and kept the sentences that have nested named entities. 
For consistency, we employ the same three annotators to further label those sentences. 
Here we use the predictions of the trained NNER model because labeling samples from scratch can result in a high mislabel rate. Instead, by providing a number of candidates of potential named entities, the annotation process can be greatly boosted with higher accuracy. The annotators will delete the wrong predictions and add the missing named entities.
Among the 50,000 sentences, 10,000 sentences with nested entities are kept, we further kept 10,000 sentences with the largest number of flat named entities to ChiNesE to offer more training samples.

Finally, to ensure an accurate metric, we select 1,000 and 1,500 sentences that have nested named entities from the 10,000 nested sentence to form our validation set and the test set. We re-train the Biaffine model based on the rest 17500 sentences and generate the prediction on the 2500 samples of the validation set and the test set. We hire the same 3 annotators to further refine the annotation of the 2,500 sentences carefully, aided by the prediction of the Biaffine model. 

Table~\ref{tab-static} gives the static of ChiNesE and the parsed ACE 2005, and Table~\ref{tab-distribution} gives the distribution their entities. Our dataset is much larger than ACE 2005 with deeper depth and more categories. The nested named entities in ChiNesE are more diverse compared to ACE 2005: 31 percent of nested entities share the last character with another entity in ChiNesE while there is only 0.05 percent of nested entities in the parsed ACE 2005 have the same last character with another entity.
Moreover, ChiNesE is specially designed for CNNER, most of the sentences in the validation set and the test set have nested named entities, which can best test the model performance on CNNER.

\begin{figure*}[t]
\centering
\includegraphics[width=0.85\textwidth]{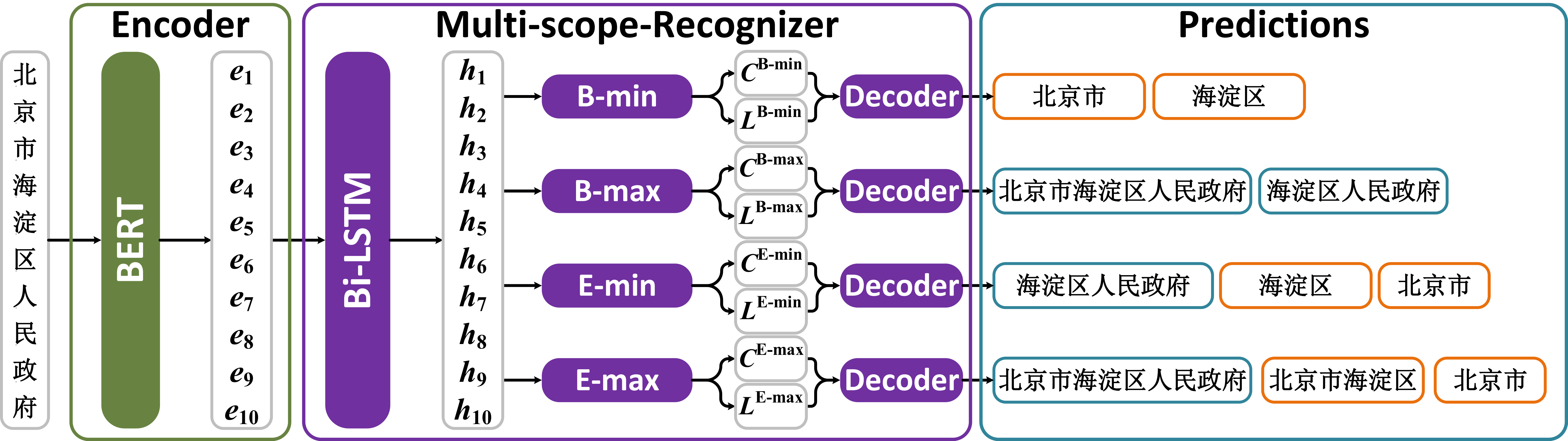} 
\caption{The structure of Mulco. The example is from Figure~\ref{fig-example}}
\label{fig-structure}
\end{figure*}




\section{Methodology}
\label{sec-method}
In this section, we introduce our proposed approach Mulco, a novel solution to CNNER problem. It can recognize named entities by adapting multiple scopes for accurate entity identification from nested structures.

\subsection{Recognition through a Scope}
The NNER task aims to recognize all entities from a given sentence. Inspired by modern Computer Vision methods \cite{jiang2022review}, we use scopes to locate the named entities in a sentence under a sequence labeling scheme, which recognizes named entities by finding their anchor and predicting its length. We define $\tt P\text{-}x$ as the scope which uses the $x^{\tt th}$ token of the named entity to locate it. For example, if we use $\tt P\text{-}1$ to locate ``Beijing Tiananmen'' in ``I am going to Beijing Tiananmen'', we first find the position of the word ``Beijing'' (first token of ``Beijing Tiananmen''), which is the $5^{\tt th}$ token of the sentence. Next, by knowing the length of the ``Beijing Tiananmen'' is $2$, we can locate the last token ``Tiananmen'', which is $6^{\tt th}$ token of the sentence. The entity ``Beijing Tiananmen'' can then be extracted from the sentence.

We employ the sequence labeling method to enable the above recognition method. Specifically, for a given sentence $T=\{t_i\}_{1\leq i\leq N}$, where $t_i$ represent the $i^{\tt th}$ character of the sentence, and $N$ is the total number of characters. In Chinese, a character acts as a token, just like a word in English. 
We use two sequences of labels to locate entities. The first sequence of labels are the anchor labels $C^{\tt P\text{-}x}=\{c^{\tt P\text{-}x}_i\}_{1\leq i\leq N}$, where $c^{\tt P\text{-}x}_i$ is the anchor label of character $t_i$ for the scope ${\tt P\text{-}x}$. 
We have
\begin{equation*}\label{equ-anchor}
c^{\tt P\text{-}x}_i=
\begin{cases}
 \tt NA, & \ t_i\ \tt{is\ not\ an\ anchor}\\
 \tt {\tt cate}, & \tt otherwise\\
\end{cases},
\end{equation*}
where $c^{\tt P\text{-}x}_i={\tt cate}$ if $t_i$ is the $x^{\tt th}$ token of an named entity, and $\tt cate$ is the category of the entity with the anchor $t_i$. $\tt NA$ means ``not an anchor''.

By predicting the anchor labels, we can locate a certain character of an entity and know its category. However, to fully recognize the named entity, we also need a sequence of the length labels $L^{\tt P\text{-}x}=\{l^{\tt P\text{-}x}_i\}_{1\leq i\leq N}$ to obtain the lengths of the entities, where $l^{\tt P\text{-}x}_i$ is the length label of $t_i$. We define
\begin{equation*}\label{equ-length}
l^{\tt P\text{-}x}_i=
\begin{cases}
 \tt 0, & \ t_i\ \tt{is\ not\ an\ anchor}\\
  z, & \tt otherwise\\
\end{cases},
\end{equation*}
where $z$ is the length of the entity which has $t_i$ as its anchor for the scope $\tt P\text{-}x$.

It is obvious that in flat NER, where no nested named entity is considered, we can recognize all named entities by locating anchors and their lengths with a scope $\tt P\text{-}x$. However, when recognizing named entities from NNER, using single scope will no longer work. For example, if we also define ``Beijing'' as a named entity of location while having ``Beijing Tiananmen'', when we use $\tt P\text{-}1$ as the scope, ``Beijing'' will then be the anchor of both ``Beijing'' and ``Beijing Tiananmen'', and that will cause critical problems on the sequence labeling. 

\subsection{Recognition through Multiple Scopes}
An intuitive way to recognize nested named entities is to employ multiple scopes to handle the nested structures between named entities. Thus, in Mulco, we use four different scopes to solve CNNER problem. 

Figure~\ref{fig-example} gives an example of how the four scopes work. For convenience, we use $\tt B\text{-}{min}$, $\tt B\text{-}{max}$, $\tt E\text{-}{min}$ and  $\tt E\text{-}{max}$ to denote the four scopes used in Mulco. $\tt B\text{-}{min}$, $\tt B\text{-}{max}$ use the first token of an entity as the anchor and predict the named entities of the shortest and the longest length, while $\tt E\text{-}{min}$ and  $\tt E\text{-}{max}$ use the last token of an entity as the anchor and recognize the named entities with the shortest and the longest lengths. The example gives a heavily nested structure of name entities. None of the four scopes can find all named entities only by itself. For example, $\tt B\text{-}{min}$ is unable to recognize ``Haidian District People's Government'', ``Haidian District, Beijing Municipality'' or ``Haidian District People's Government of Beijing Municipality''. On the other hand, $\tt B\text{-}{max}$ failed to recognize ``Beijing Municipality'', ``Haidian District'' and ``Haidian District, Beijing Municipality''. However, if we combine the two scopes, we can cover most entities in the example except ``Haidian District, Beijing Municipality'', which is later covered by $\tt E\text{-}{max}$.
By aggregating the results of all the four scopes, Mulco is able to decode the name entity in any nested structures (100\% of entities in ChiNesE).

\subsection{Mulco}

With the effective scope-based recognizing method designed above, we are able to solve the CNNER problem in sequence labeling approach.
Figure~\ref{fig-structure} illustrates the structure of Mulco. Given a sentence $T=\{t_i\}_{1\leq i\leq N}$, we first encode the sentence with 
\begin{equation*}\label{equ-bert}
\boldsymbol{E}=\{\boldsymbol{e}_i\}_{1\leq i\leq N}=BERT(T),
\end{equation*}
where $\boldsymbol{E}$ is the sequence of the character embeddings of $T$, $\boldsymbol{e}_i \in \mathbb{R}^d$, $d$ is the dimension of the embedding and $BERT(\cdot)$ extract the BERT embedding of the input sentence.

Since we use a single character to represent the anchor of an entity, the contextual information will be very important for better recognition. Although BERT can aggregate the contextual information of a character using its self-attention mechanic, a Bi-LSTM structure may further improve it. Thus, we employ a multi-layer Bi-LSTM to calculate the contextualized embedding $\boldsymbol{H}$ by
\begin{equation*}\label{equ-lstm}
\boldsymbol{H}=\{\boldsymbol{h}_i\}_{1\leq i\leq N}=Bi{\text -}LSTM(\boldsymbol{E}),
\end{equation*}
where $\boldsymbol{h}_i \in \mathbb{R}^m$, $m$ is the dimension of the contextualized embedding and $Bi{\text -}LSTM(\cdot)$ extract the embeddings from the inputs.

Once obtained the contextualized embedding, in each scope, we employ two linear classifiers to predict the anchor label and the length label for $t_i$. Here we take $\tt B\text{-}{min}$ as an example:
\begin{equation}\label{equ-loss1}
\begin{split}
\boldsymbol{\boldsymbol{c}}^{\tt B\text{-}min}_i&= \boldsymbol{W}^{\tt B\text{-}min}_C\boldsymbol{h}_i + \boldsymbol{b}^{\tt B\text{-}min}_C,\\
\boldsymbol{\boldsymbol{l}}^{\tt B\text{-}min}_i&= \boldsymbol{W}^{\tt B\text{-}min}_L\boldsymbol{h}_i + \boldsymbol{b}^{\tt B\text{-}min}_L,\\
\end{split}
\end{equation}
where $\boldsymbol{{c}^{\tt B\text{-}min}_i} \in \mathbb{R}^{N_c}$ and $\boldsymbol{{l}^{\tt B\text{-}min}_i}\in \mathbb{R}^{N_l}$ are the $i^{\tt th}$ (one-hot) anchor label and the (one-hot) length label of $\tt B\text{-}{min}$ respectively. 
$\boldsymbol{W}^{\tt B\text{-}min}_C\in\mathbb{R}^{m\times N_c}$, $\boldsymbol{b}^{\tt B\text{-}min}_C\in\mathbb{R}^{N_c}$, $\boldsymbol{W}^{\tt B\text{-}min}_L\in\mathbb{R}^{m\times N_l}$ and $\boldsymbol{b}^{\tt B\text{-}min}_L\in\mathbb{R}^{N_l}$ are the trainable parameters of the linear classifiers of $\tt B\text{-}{min}$. $N_c$ represents the number of categories and $N_l$ is a hyper-parameter represents the max length of all entities.

For $\tt B\text{-}{min}$, denote the predicted sequence of anchor labels as $C^{\tt B\text{-}min}=\{\boldsymbol{c}_i^{\tt B\text{-}min}\}_{1\leq i\leq N}$, and the predicted sequence of length labels as $L^{\tt B\text{-}min}=\{\boldsymbol{l}_i^{\tt B\text{-}min}\}_{1\leq i\leq N}$. By employing the cross-entropy function, the training object for $\tt B\text{-}{min}$ is then
\begin{equation}\label{equ-loss2}
\begin{split}
    \mathcal{L^{\tt B\text{-}min}}=\sum_{i=1}^{N}{\ell_i^{\tt B\text{-}min}},
\end{split}
\end{equation}
where 
\begin{equation}\label{equ-loss3}
\begin{split}
    {\ell_i^{\tt B\text{-}min}}=-\sum_{j=1}^{N_c} \bar{{c}}^{\tt B\text{-}min}_{i,j}\log{c}^{\tt B\text{-}min}_{i,j} -\sum_{k=1}^{N_l}\bar{{l}}^{\tt B\text{-}min}_{i,k}\log{l}^{\tt B\text{-}min}_{i,k}.
\end{split}
\end{equation}
Here $c^{\tt B\text{-}min}_{i,j}$ and $l^{\tt B\text{-}min}_{i,k}$ are the $j^{\tt th}$ and the $k^{\tt th}$ elements of ${c}^{\tt B\text{-}min}_{i}$ and ${l}^{\tt B\text{-}min}_{i}$ respectively, 
while $\bar{c}^{\tt B\text{-}min}_{i,j}$ and $\bar{l}^{\tt B\text{-}min}_{i,k}$ are their corresponding true labels. 

The overall training object is then
\begin{equation*}
\mathcal L = \mathcal{L^{\tt B\text{-}min}}+\mathcal{L^{\tt B\text{-}max}}+\mathcal{L^{\tt E\text{-}min}}+\mathcal{L^{\tt E\text{-}max}},
\end{equation*}
where $\mathcal{L^{\tt B\text{-}max}}$, $\mathcal{L^{\tt E\text{-}min}}$ and $\mathcal{L^{\tt E\text{-}max}}$ are the training objects of the other scopes, obtained with the same method following Equation~(\ref{equ-loss1})-(\ref{equ-loss3}).

As shown in the Figure~\ref{fig-structure}, we extract the predicted named entities of all scopes as the results. We will keep the category with the highest confidence if an entity is recognized by multiple scopes with different categories.
   
    

\section{Experiments}
\label{exp}
To investigate the effectiveness of Mulco, we evaluate our proposed method on ChiNesE and ACE 2005, together with other state-of-the-art single-model methods that adapt different NNER approaches. Following previous literature on NNER, we report standard precision (P), recall (R) and micro F1-score (F1) to evaluate the performance.

\subsection{Data}
Besides ChiNesE, we also use ACE 2005 \cite{doddington-etal-2004-automatic} to further demonstrate the performance of all baselines.
ACE 2005 is a corpus collected from broadcasts, newswires, and weblogs in Arabic, Chinese and English. It is one of the most popular datasets to develop NNER methods.
However, though some research is done for CNNER using ACE 2005 Chinese corpus, there is no commonly used division for training, validation, and testing. We follow the setting of \cite{lu2015joint, ju2018neural} and parse the Chinese corpus by ourselves. Specifically, we use 8:1:1 to divide all documents in ACE 2005 into the training set, validation set, and test set. We then split the documents into sentences to form the samples. The detailed static are shown in Table~\ref{tab-static} and Table~\ref{tab-distribution}. 

\begin{table*}
	\centering
	\small
	\begin{tabular}{l c c c c c c c}
	\toprule 
	\multirow{2}{*}{} & \multirow{2}[2]{*}{Approach} &\multicolumn{3}{c}{ACE 2005}&\multicolumn{3}{c}{ChiNesE}\\ \cmidrule(l){3-5}\cmidrule(l){6-8}
	&& P & R  & F1 & P & R  & F1\\
	\midrule
	    Innermost & Sequence Labeling
	    &89.27&65.74&75.72
	    &93.79&54.42&68.88\\
	    Outermost & Sequence Labeling
	    &\textbf{89.38}&61.54&72.83
	    &\textbf{94.21}&43.94&59.93\\
	    \citet{wang-etal-2018-neural-transition} &Transition-based
	    &77.62&81.60&79.56
	    &86.56&76.34&81.13\\
	    \citet{wang-etal-2020-pyramid} & Layer-based 
	    &83.22&76.29&84.72
	    &90.79&77.71&84.02\\
	    \citet{yu-etal-2020-named} &Span-based 
	    &81.57&86.63&84.02
	    &92.14&77.69&84.30\\
	    \midrule
	    Mulco & Sequence Labeling
	    &83.89&\textbf{88.17}&\textbf{85.98}
	    &90.57&\textbf{80.47}&\textbf{85.23}\\
	    \bottomrule 
	\end{tabular}
	\caption{The experimental results on ChiNesE and ACE 2005.}
	\label{tab-result}
\end{table*}

\subsection{Baseline Methods}
In Table~\ref{tab-result}, Innermost and Outermost use traditional ``BIOES'' sequence labeling method \cite{10.1109/TKDE.2020.2981314, liu2022chinese} and only recognize the shortest and the longest named entities. 
We also test three state-of-the-art models which are designed based on different labeling method. Specifically, Pyramid is a layer-based method designed by stacking inter-connected layers to predicts whether a text segment with a certain length belongs to an entity. The length of the text segment (n-gram) to be detected for each layer increases with the depth of the layer, which forms a pyramid shape in the model structure. 
\citet{wang-etal-2018-neural-transition} propose a scalable transition-based method, where each sentence is transformed into a tree consisting of its words. They design actions to represent the nest structures of entities, treats the words in a sentence as buffers and tries to reconstruct the tree by predicting a sequence of actions. 
Biaffine \cite{yu-etal-2020-named} is a span-based method. It predicts all possible regions in a sentence to find entities. For each region, a representation of the start token and the end token is generated to represent the information of the whole text piece of the region when predicting.


\subsection{Experimental Setting}
All experiments are done on a single NVIDIA RTX 3090 GPU.
For a fair comparison, we fine-tune BERT based on ``bert-base-chinese''\footnote{https://huggingface.co/bert-base-chinese} and only use the output BERT embedding as the inputs for all models. 

For \textbf{Mulco}, we use 4-layer and 2-layer Bi-LSTM for ACE 2005 and ChiNesE respectively. The dimension of the contextualized embedding is set to 1536. We train Mulco for 50 epochs with a batch size of 16 on both datasets. For ACE 2005 and ChiNesE, the learning rate of BERT is 0.00002 and 0.00001 while the learning rate of the other trainable parameters is 0.0002 and 0.0001. The drop-out rates are 0.5 and 0.3 respectively and the weight decay factors are 0.05 for both datasets. We set the max entity length to 512, same as the max length of input text. We use AdamW \cite{loshchilov2018decoupled} as the optimizer.
For \textbf{Innermost and Outermost}, we train them for 50 epochs with a batch size of 16 on both datasets. The learning rates are set to 2e-5, with a drop-out rate of 0.5 and a weight decay factor of 0.05.
For \textbf{other baseline methods}, we fine-tune BERT with a learning rate of 0.00001. We train Biaffine for 100 epochs and the other baselines for 150 epochs to ensure convergence. Other parameters are the same as their original experimental settings.

\subsection{Results}
Table~\ref{tab-result} gives the results of all experiments. As expected, the Innermost and the Outermost methods reach the highest precision on both datasets with their traditional sequence labeling approach. However, such a labeling scheme can not handle the nested named entities, thus resulting in much lower recall rates compared with other baselines. The recall rates of Innermost are much higher than the recall rates of Outermost because when nested, there could be multiple short named entities in a long named entity.

The transition-base method does not perform well on both datasets. It generates actions to decode nested named entities and uses special labels to link two adjacent words belonging to the same entity. It works fine in English NNER since the named entities usually consist of a few words. However, in CNNER, each named entities usually consist of many characters, which brings extra challenges to the model. 

Pyramid and Biaffine perform similarly on both datasets. Pyramid performs better when the ratio of the nested named entities is not very high (in the test set of ACE 2005), while Biaffine, as a region-based method, performs better in the situation where the entities are heavily nested (in the test set of ChiNesE).

Our proposed method achieved the best overall performance (F1) on both datasets. It outperforms other models by at least 1.26 on ACE 2005 and 0.93 on ChiNesE.
By employing multiple scopes, Mulco has higher recall rates than all other baseline methods on both datasets which are 1.54\% and 2.76\% higher than all other models.
This is because each scope is specially designed for a certain type of named entities, by gathering the predictions of all scopes, more potential named entities can be retrieved from the given sentence. Although such a mechanic will lead to a lower precision, the overall performance is greatly boosted. Also, since NER is one of the first steps of information retrieval, a higher recall rate may benefit other following downstream tasks.

\begin{table}
\small
	\centering
	\begin{tabular}{l c c c c c c}
	\toprule
	\multirow{2}{*}{}  &\multicolumn{3}{c}{ACE 2005}&\multicolumn{3}{c}{ChiNesE}\\ \cmidrule(l){2-4}\cmidrule(l){5-7}
	& P & R  & F1 & P & R  & F1\\
	\midrule
	    Mulco
	    &83.89&\textbf{88.17}&85.98
	    &90.57&\textbf{80.47}&\textbf{85.23}\\
	    \midrule
	    $\tt B\text{-}min$
	    &\textbf{86.87}&69.41&77.16
	    &92.11&57.44&70.75\\
	    $\tt B\text{-}max$
	    &84.66&67.60&75.17
	    &91.12&56.85&70.02\\
	    $\tt E\text{-}min$
	    &86.19&87.73&\textbf{86.95}
	    &93.14&72.13&81.30\\
	    $\tt E\text{-}max$
	    &86.20&87.59&86.89
	    &\textbf{93.19}&72.05&81.27\\\midrule
	    -LSTM
	    &82.20&87.15&84.60
	    &89.71&80.71&84.98\\
	    \bottomrule 
	\end{tabular}
	\caption{The results of the ablation studies.}
	\label{tab-ablation}
\end{table}
\subsection{Ablation Studies}
\label{sec-ablation}
We conduct the ablation studies to verify the effectiveness of the components of Mulco and report the experimental results on both datasets in Table~\ref{tab-ablation}.

\textbf{Scope v.s. traditional sequence labeling.} We first examine the performance of each scope of Mulco. $\tt B\text{-}min$ and $\tt B\text{-}max$ have higher recall rates but lower precision than the Innermost and Outermost methods. Although they all use sequence labeling and aim to find the longest and the shortest named entities, they recognize named entities with different methods. The Innermost and the Outermost methods use the ``BIOES'' scheme to label each token in a sentence. 
To recognize an entity, each token in the entity needs to be correctly classified. Since any wrong prediction will fail to retrieve the entity, the precision of those two methods is much higher than all other methods. On the other hand, $\tt B\text{-}min$ and $\tt B\text{-}max$ predict an entity by recognizing its first token and the length of the entity. Each scope uses two classifiers to recognize the entities, which lowers the precision. However, such an approach brings more advantages to recalling more candidates, it can retrieve the entities that the traditional ``BIOES'' labeling method can not recognize, which results in a higher recall rate and overall performance.

Take $\tt B\text{-}max$ in Figure~\ref{fig-example} as an example, it can properly recognize ``Haidian District People's Government of Beijing Municipality'' and ``Haidian District People's Government'' with a single scope: $\tt B\text{-}max$ recognizes an entity only by the anchor label and the length label of its first token.
The two entities have two different start characters, thus can be distinguished by $\tt B\text{-}max$.
In contrast, the Outermost method can only recognize ``Haidian District People's Government of Beijing Municipality'', since the traditional sequence labeling method considers all characters in an entity when recognizing thus only allowing one character to have one label even it is an element of more than one entities.

\textbf{Diversity of nested named entities.} We also have some interesting funding on the experiment results of $\tt E\text{-}min$ and $\tt E\text{-}max$. As shown in Table~\ref{tab-ablation}, $\tt E\text{-}min$ and $\tt E\text{-}max$ have higher performance than $\tt B\text{-}min$ and $\tt B\text{-}max$ on both datasets, they even achieve higher performance than the default Mulco scope setting on ACE 2005 with only single scope. This is because, in Chinese, nested named entities are more likely to share the same start characters with other named entities. Among 1801 nested named entities in the test set of ACE 2005, 1479 of them are in such a situation, while there are only 4 named entities that have the same end characters as other entities. We can see that the ratio (1479:4) is extremely unbalanced. This is why a single scope using the end character as the anchor can achieve even higher performance than the combination of four scopes. However, ChiNesE has a more balanced ratio, which is 7371:3218. Our dataset is collected from online passages of multiple domains, thus being more diverse and closer to the real-word CNNER application. Although $\tt E\text{-}min$ and $\tt E\text{-}max$ still outperforms $\tt B\text{-}min$ and $\tt B\text{-}max$ on ChiNesE, the portion of named entities they can cover is much smaller than it on ACE 2005. By combining the four scopes, Mulco reaches a much higher performance than the performance of any single scope on ChiNesE.

\textbf{Necessity of contextual information.}
In Table~\ref{tab-ablation}, ``-LSTM'' reports the performance of Mulco without Bi-LSTM encoder on ACE 2005 and ChiNesE. The F1 drops 1.38 on ACE 2005 and 0.25 on ChiNesE. The drop on ChiNesE is much lower than it on ACE 2005. This is because BERT, as a transformer-based pre-trained language model, also can grab contextual information for an input character. With a large enough dataset, BERT can better understand how to pay attention to the contextual information for the CNNER task. Thus, Mulco can have enough contextual information for prediction even without LSTM on ChiNesE. However, on ACE 2005, the data size can not sufficiently support BERT to learn how to extract the contextual information, where a Bi-LSTM model is in need to enhance such an ability. 
Moreover, the recognition of scope depends on the embedding of a single character, thus including the contextual information into the embeddings becomes critical to the performance of our model. If we use a pre-trained word vector instead of BERT embedding as input, a contextualization model such as LSTM will then become necessary.

\section{Conclusion}
In this paper, to facilitate research on Chinese Nested Named Entity Recognition (CNNER), 
we construct ChiNesE, a CNNER dataset containing 20,000 high-quality samples with 117,284 named entities falling into 10 categories, where 43.8 percent of them are nested. ChiNesE are sampled from online passages of multiple domains, which have more diverse nested named entities and are closer to the real-world CNNER application today.
To fill the gaps in sequence labeling and CNNER, we propose a scope-based labeling method, where each scope extracts named entities by predicting the anchor and the length of an entity. It can avoid the limitation of the traditional sequence labeling scheme. We then propose Mulco, a novel method that utilizes four different scopes to effectively recognize nested named entities. Experimental results show that Mulco has achieved the best performance on ChiNesE and ACE 2005.

\bibliography{aaai23}
\end{document}